\documentclass{article}
\pdfoutput=1

    \usepackage[preprint, nonatbib]{neurips_2020}

\usepackage[utf8]{inputenc} 
\usepackage[T1]{fontenc}    
\usepackage{url}            
\usepackage{booktabs}       
\usepackage{amsfonts}       
\usepackage{nicefrac}       
\usepackage{microtype}      

\usepackage{graphicx}
\usepackage{bmpsize}
\usepackage{subcaption}
\usepackage{float}
\usepackage{url}
\usepackage{amsmath}
\usepackage{bm}
\usepackage{caption}
\title{Dual Encoding U-Net for Spatio-Temporal Domain Shift Frame Prediction 
}

\author{Jay Santokhi\hspace{3mm}Dylan Hillier\hspace{3mm}Yiming Yang\hspace{3mm}Joned Sarwar\hspace{3mm}Anna Jordan\hspace{3mm}Emil Hewage\\[1mm]
      Alchera Data Technologies Ltd \\
      Cambridge, CB2 1NN\\
      \texttt{\{jay,dylan,yiming,joned,anna,emil\}@alcheratechnologies.com} 
   
}

\begin{document}

\maketitle

\begin{abstract}
The landscape of city-wide mobility behaviour has altered significantly over the past 18 months. The ability to make accurate and reliable predictions on such behaviour has likewise changed drastically with COVID-19 measures impacting how populations across the world interact with the different facets of mobility. This raises the question: "How does one use an abundance of pre-covid mobility data to make predictions on future behaviour in a present/post-covid environment?" This paper seeks to address this question by introducing an approach for traffic frame prediction using a lightweight Dual-Encoding U-Net built using only 12 Convolutional layers that incorporates a novel approach to skip-connections between Convolutional LSTM layers. This approach combined with an intuitive handling of training data can model both a temporal and spatio-temporal domain shift (\url{gitlab.com/alchera/alchera-traffic4cast-2021}). 

\end{abstract}
\section{Introduction}
The traffic frame prediction challenge, Traffic4Cast 2021, presented by the Institute of Advanced Research in Artificial Intelligence has changed from the prior iterations in 2019 \cite{KoppMichaelKopp2020} and 2020 \cite{Kopp2021}. The training and testing data distributions in previous iterations had a degree of overlap. However, the 2021 edition poses a few-shot learning problem \cite{Wang2020} where training data is limited and varies considerably from testing data requiring models that can adapt to domain shifts in time and space.

Despite the changes in the 2021 edition, previous approaches still have merit. For both the 2019 and 2020 iterations, models that incorporated any similarity to U-Nets \cite{Ronneberger2015} seemed to outperform those without \cite{Chao2018, Choi2020, Herruzo2019, Martin2019}, however there has been an emergence in Graph Neural Network based models \cite{Qi2020, Maas2020}. In order to apply and further the potential of U-Net style methods the training process needs to become more sophisticated. This can be accomplished through a variety of ways such as Transfer Learning, Fine-Tuning and Pre-Training \cite{Xu2020}. A combination of such approaches could mitigate the gap between the training and testing data enabling a reliable set of predictions.

The necessity for such a challenge is increasingly apparent as the impact of Covid-19 on mobility becomes clearer. Taking London as an example and looking at Alchera Technologies' Mobility Dashboard \cite{AlcheraDataTechnologiesLtd} it can be clearly seen how covid measures have reduced the usage of cars and buses on a city wide scale during the pandemic. Covid aside, cities are dynamic, complex systems and not topologically static and with these factors combined prediction systems can quickly become outdated. This makes a lightweight, quick to train model with competitive performance incredibly desirable. 
Given these points, this paper aims to iterate on \cite{Santokhi2020}'s work from Traffic4Cast 2020 and showcase a model architecture that utilises Convolutional-LSTMs \cite{Shi2015} to form a lightweight Dual Encoding U-Net with fewer than 500k parameters that embraces a novel approach to skip connections along with an intuitive training approach that incorporates pre-training, fine-tuning and ensembling to make predictions for both core (Temporal Domain Shift) and extended (Spatio-Temporal Domain Shift) challenges.

\section{Data and Problem Definition}
The dataset provided for the two tasks consists of 10 different cities from around the world over the course of 2 years. The cities have been split into 3 categories: Core Challenge, Extended Challenge and Additional Training. The core challenge data consist of 4 cities: Berlin, Chicago, Istanbul and Melbourne. The extended challenge data consists of 2 cities: New York and Vienna, and the additional training data consists of another 4 cities: Antwerp, Bangkok, Barcelona and Moscow. 

\begin{figure}[ht]
  \centering
  \includegraphics[width=0.875\linewidth]{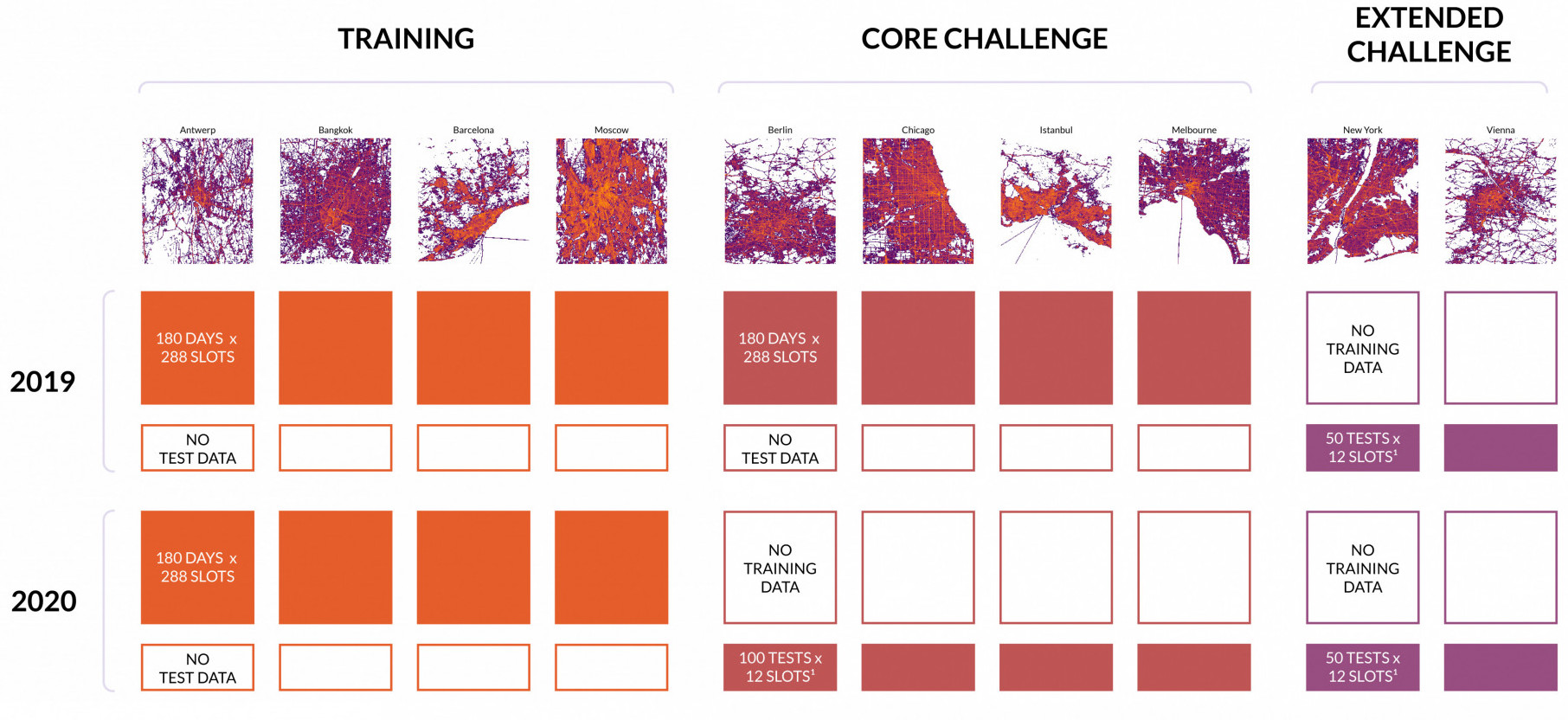}
  \small \caption{\scriptsize Breakdown of competition data (further details found at \cite{github})}
  \label{fig:data}
\end{figure}

The area of each city is presented as an image frame of size 495x436 where each pixel represents a region of 100m$^{2}$. The time bins between each frame is 5 minutes therefore one day of data is given as a tensor of size (288, 495, 436, 8). The dynamic information encoded in the 8 channels consists of the traffic volume and average speed per heading direction: NE, SE, SW, or NW e.g. channel 1 and 2 contain the volume and speed of vehicles heading NE respectively. Figure \ref{fig:data} provides a break down of the data available for each of the cities.
\vspace{-3mm}
\vspace{-6mm}
\subsection{Temporal Domain Shift Task}
\begin{figure}[ht]
  \centering
  \includegraphics[width=0.675\linewidth]{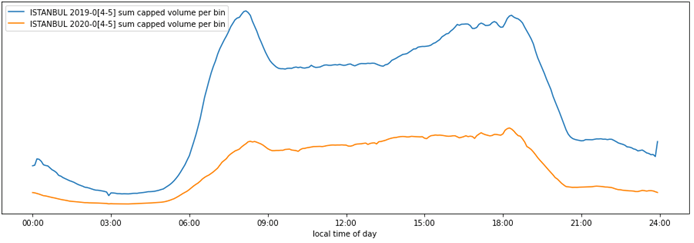}
  \small \caption{\scriptsize Illustration of temporal domain shift due to pandemic. Traffic volume (sum of all directions)
in a 24h interval in 2019 (blue) and 2020 (orange) in Istanbul \cite{iarai}}
  \label{fig:shift}
\end{figure}

For the Core Challenge, the task is to make predictions on the traffic states 5, 10, 15, 30, 45 and 60 minutes into the future for each of the 4 cities when given 100 1-hour test samples in 2020, however only 6 months of 2019 training data is provided for each of the cities. Therefore, any model developed must be able to handle the temporal shift between pre-covid (2019) and present-covid (2020). Figure \ref{fig:shift} highlights the shift between 2019 and 2020 data.

\vspace{-4mm}

\subsection{Spatio-Temporal Domain Shift Task}
For the extended challenge, the task remains similar - using the previous hour (12 frames) to make predictions on the next 5, 10, 15, 30, 45 and 60 minutes, however the 100 test slots contain 50 samples from 2019 and 50 samples from 2020 with no labels on which sample is from which year. In addition to this, the cities for which predictions are to be made do not have any training data. 

Therefore, any model developed not only has to adjust between domain shifts in time but also has to adjust to a shift in space by making predictions on entirely unseen cities.

\section{Method}
\subsection{Data Preparation}
Given the nature of the data for these frame prediction tasks there are 3 main steps that need to be taken into consideration during a data preparation phase: pre-processing, sampling strategy and output sequence length. 

Given the raw data values for both speed and volume have been encoded to be between 0-255 (0 referring to be zero volume or speed lower than minimum and 1-255 denoting a linear interpolation of real values up to a maximum) normalisation of values to be between 0-1 was completed during the pre-processing step.

For the sampling strategy step, two approaches are available to obtain \textit{(data, label)} training pairs: overlapping (with varying window sizes) or non-overlapping. Opting for a non-overlapping approach will yield 2172 sequence pairs per city in comparison to an overlapping strategy with a sliding window of 1 which will yield a total of 47965 sequence pairs per city. For the purpose of this work a non-overlapping approach was predominately used to allow shorter training cycles aiding in selection of hyperparameters. Once optimal parameters were identified an overlapping approach with a window size of 12 (4163 sequence pairs) was adopted to ensure samples from each hour of the day were present in the training data.

The last step in the data preparation is selecting the output length. Using the previous hour to predict the next hour implies using 12 frames to predict the next 12 frames, given the granularity between each frame is 5 minutes. The problem definition however is only interested in 6 of these frames (5, 10, 15, 30, 45, 60 minutes) resulting in 6 predicted frames being thrown away. If one was to consider deployment of such models in a practical environment a consistent granularity of predictions would be more desirable to users and road operators. In spite of that, for maximising core competition performance it has been found that forcing the models to learn just the 6 frames of interest has yielded better results than predicting all 12 frames in an hour and removing the ones not of interest. 

\subsection{Model Architecture}
\begin{figure}[ht]
  \centering
  \includegraphics[width=0.87\linewidth]{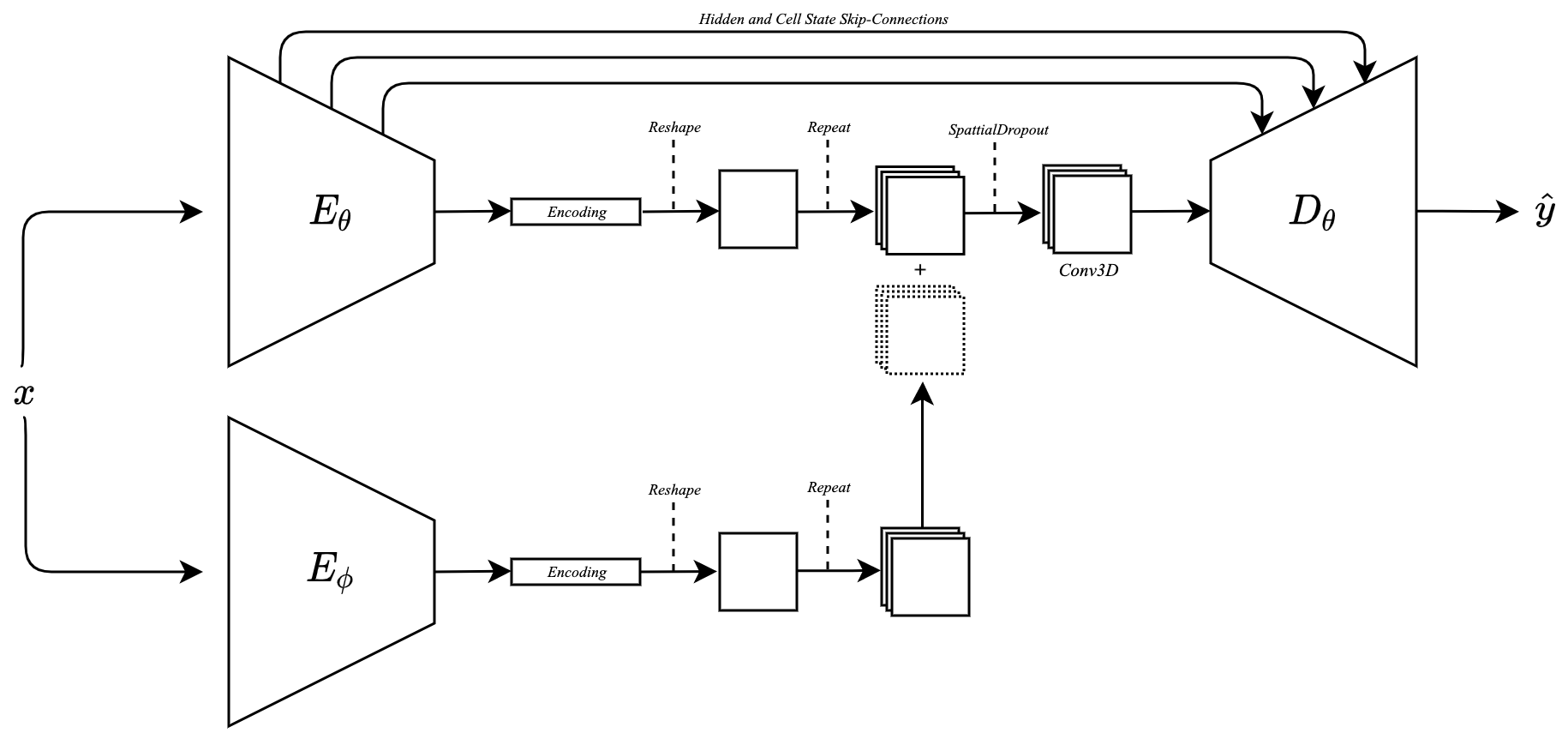}
\vspace{-1mm}
  \scriptsize\caption{\scriptsize Final Model Architecture: $E_\theta$ and $E_\phi$ each contain 3 ConvLSTM layers and 2 Max-Pooling layers. $D_\theta$ contains 3 ConvLSTM layers and 2 Transposed Convolution layers. Skip-connections carry hidden and cell states of ConvLSTMs in $E_\theta$ to use as initial states for ConvLSTMs in $D_\theta$. The encodings learnt from $E_\theta$ and $E_\phi$ are reshaped to a 2D tensor before being repeated. An Addition layer is used to combine these prior to Spatial Dropout and a 1x1 3D Convolution layer. $D_\theta$ then upscales to form the required predictions. The core challenge model uses more filters in the ConvLSTM layers than the extended challenge model. The extended challenge model also does not contain the Spatial Dropout or the 3D Convolution. The result of the Addition layer is fed straight into $D_\theta$ in this case. Mean Squared Error loss was used for training and validation in both challenges.
  }
  \label{fig:model}
\end{figure}

The model architecture developed for the two tasks consists of an Autoencoder structure that adopts two encoders with skip connections between one of the encoders and the decoder. This Dual-Encoding U-Net however incorporates a novel approach to its skip connections making them more appropriate and intuitive to the task of frame prediction; vanilla U-Nets were originally developed for image segmentation utilising 10s of 2D Convolution layers whereas the model shown here is built using 9 Convolutional LSTM layers thus requiring a different approach to skip connections. The following sections will delve into the iterative steps used to reach the final architecture seen in Figure \ref{fig:model}.
\vspace{-3mm}


\subsubsection{Rethinking Skip-Connections}
Starting with a baseline model that includes Encoder, $E_{\theta}$ and Decoder, $D_{\theta}$, simple skip-connections between $E_{\theta}$ and $D_{\theta}$ are implemented using an Addition layer to combine the outputs of respective Convolutional LSTM layers betweeen $E_{\theta}$ and $D_{\theta}$:
\begin{equation}
    \mathcal{O}^{i}, \mathcal{H}^{i}, \mathcal{C}^{i} = f^{i}_{D_\theta}\left(\mathcal{O}^{i-1} + \mathcal{O}^{j}, \emptyset, \emptyset \right)
\end{equation}
Where $f^{i}_{D_\theta}$ denotes the operations of a Convolutional LSTM in the $i$th layer of ${D_\theta}$ and $\emptyset$ denotes an empty set. The output, $\mathcal{O}^i$, hidden state, $\mathcal{H}^i$ and cell state, $\mathcal{C}^i$ matrices are  returned by $f^{i}_{D_\theta}$ where the inputs were the summation of the preceding layer in ${D_\theta}$, $\mathcal{O}^{i-1}$ and the output of a corresponding $j$th layer in $E_\theta$, $\mathcal{O}^j$.

The effect of this is to ensure that information distinguishing pixels in $E_\theta$ is not lost through the max-pooling layers and can be recombined with the information processed in $D_\theta$. The issue with this standard approach is that the temporal dimension of the frames in $E_\theta$ are intended to encode the previous hour, whereas the temporal dimension in $D_\theta$ is intended to represent the following/predicted hour. Summation of these two temporal dimensions via simple addition subsequently hinders the ability of $D_\theta$ to learn the features of the following hour as encodings of frames from the preceding hour are summed to the inputs of $f^{i}_{D_\theta}$. As such two new methods for altering the skip-connections were trialed, so that the previous hours information could be disentangled more easily.

The first method, Temporal-Concatenation skip-connections, passed an entire sequence from the $j$th layer of $E_\theta$ and fed it into $f^{i}_{D_\theta}$ to obtain updated hidden and cell states, $\mathcal{H}^{i*}$, $\mathcal{C}^{i*}$ (see Equation \ref{eq:temporal_skip_1}) before feeding in the output of the previous layer in $D_\theta$, $\mathcal{O}^{i-1}$ using $\mathcal{H}^{i*}$ and $\mathcal{C}^{i*}$ as initial states for $f^{i}_{D_\theta}$ (see Equation \ref{eq:temporal_skip_2}). This is equivalent to a concatenation in the temporal dimension.
\begin{equation}
    \mathcal{O}^{i*}, \mathcal{H}^{i*}, \mathcal{C}^{i*} = f^{i}_{D_\theta}(\mathcal{O}^j, \emptyset, \emptyset)
    \label{eq:temporal_skip_1}
\end{equation}    
\begin{equation}    
    \mathcal{O}^{i}, \mathcal{H}^{i}, \mathcal{C}^{i} = f^i_{D_\theta}(\mathcal{O}^{i-1}, \mathcal{H}^{i*}, \mathcal{C}^{i*})
    \label{eq:temporal_skip_2}
\end{equation}
To elaborate, if a ConvLSTM layer, $e$ in $E{_\theta}$ has a sequence $\mathcal{O}^{j}=t^{0}_e, ..., t^{N}_e$ and this was passed to a ConvLSTM layer, $d$ in $D{_\theta}$ via a skip-connection the resulting input sequence to $d$ is first $\mathcal{O}^{j}$ (\ref{eq:temporal_skip_1}) and the hidden and cell states learnt here  are then used as initial states when inputting $\mathcal{O}^{i-1}=t^{0}_d, ..., t^{N}_d$ (\ref{eq:temporal_skip_2}). The relationship between each $t^{i}_e$ is different to the relationship between each $t^{i}_d$. 
One potential issue with this approach is that the relationship between the sequences at the input to $E{_\theta}$ layers will not be the same as the relationship intended to be learnt between the sequences being inputted to the $D{_\theta}$ layers. By feeding in $\mathcal{O}^{j}$ followed by $\mathcal{O}^{i-1}$ with $\mathcal{H}^{i*}$ and $\mathcal{C}^{i*}$ we are forcing the ConvLSTM layer to perform both encoding and decoding.

The next method, Hidden and Cell State Skip Connections was to instead pass the last hidden and cell states from the $j$th layer in $E_{\theta}$, $\mathcal{H}^{j}$ and $\mathcal{C}^{j}$, to use as initial states for corresponding $i$th layers in $D_{\theta}$. This was inspired by the forecasting structure in the original Convolutional LSTM paper \cite{Shi2015}.
\begin{equation}
    \mathcal{O}^{i}, \mathcal{H}^{i}, \mathcal{C}^{i} = f^i_{D_\theta}(\mathcal{O}^{i-1}, \mathcal{H}^{j}, \mathcal{C}^{j})
\end{equation}
This reduces the time complexity of training $D_{\theta}$ by half compared to the Temporal-Concatenation Skip Connection method addresses the intuition regarding the relationship between temporal steps.
\vspace{-2mm}
\begin{table}[H]
  \caption{Skip-Connections}
  \centering
  \begin{tabular}{lll}
    \toprule
    \cmidrule(r){1-2}
    Type       & Training Loss &     Validation Loss \\
    \midrule
    Addition & $1.46267 \times 10^{-3}$  & $1.65217 \times 10^{-3}$ \\ 
    Sequence & $1.38826\times 10^{-3}$  & \textbf{1.54358} $\times \mathbf{10^{-3}}$ \\
    Hidden and Cell State & \textbf{1.38822} $\times \mathbf{10^{-3}}$  & $1.54403\times 10^{-3}$ \\                               
    \bottomrule
  \end{tabular}
  \caption*{\scriptsize Model trained on Moscow 2019 and validated on Moscow 2020 simulating Core Challenge}
  \label{table:skip_connection}
\end{table}
\vspace{-2mm}
From Table \ref{table:skip_connection} it can be seen that these updated skip-connections markedly improve the performance of the Addition approach when looking at the training and validation MSE loss. There is not a great distinction in training and validation loss between the two proposed skip connections, however due to improved training speed the Hidden and Cell State Skip Connections were chosen.



\subsubsection{Balance Between Pre-Training and Fine Tuning}
Given that for both the core and extended challenges the difference between training and testing data is vast, or completely lacks thereof, it is nigh on impossible to develop a prediction model without leveraging information from different cities. One way of addressing this is to carry out pre-training to warm-up model weights prior to a fine tuning step that focuses on a specific target city. This gives a more representative starting point for model weights than random initialisation as city features will have already started to be learnt. In order to optimally utilise the additional data provided a variety of `pre-training followed by fine-tuning' scenarios were attempted: using no pre-training, using only 2019 data, using only 2020 data and using both 2019 and 2020 data. 

It is safe to assume that for the extended challenge, where predictions are to be made on both 2019 and 2020, pre-training on both 2019 and 2020 data would prove to be the best approach. For the core challenge however, it was not expected that this same approach would also provide optimal results. It was assumed that pre-training exclusively on 2020 data would yield optimal results given the core challenge task requires predictions only to be made in 2020. Pre-training on both 2019 and 2020 data led to slightly improved performance in every simulated core challenge train no matter which city combinations were chosen. A potential reason for such behaviour could be that a temporal shift between 2019 and 2020 may only be seen in densely populated areas and that outskirts of certain cities may have experienced comparable values between 2019 and 2020. This would allow models to learn where to focus on with regards to locations that temporal shift is most present. This line of thought however needs to be investigated and validated further.

After the pre-training phase (using 2019 and 2020 data of the additional cities) a fine-tuning phase was started. This involved resuming training with different data, that of a specific target city. The purpose of the fine-tuning is to adjust the pre-trained weights to focus on the topology of a specific target city. Seeing as training data for the target city is from 2019 and the model needs to make predictions in 2020 a balance must be struck on the number of epochs to train for so that topological and city specific features can be learnt but not so long that the 2019 features are also captured. It was found empirically that 5 epochs was the optimal amount.
\vspace{-5mm}
\begin{table}[h]
  \caption{Pre-Training}
  \centering
  \begin{tabular}{lll}
    \toprule
    \cmidrule(r){1-2}
    Type       & Training Loss &     Validation Loss \\
    \midrule
    No Pre-Training & $ 1.38823 \times 10^{-3}$  & $ 1.54403 \times 10^{-3}$ \\ 
    Pre-Train on 2019 & $ 1.41399 \times 10^{-3}$  &  $1.46532\times 10^{-3}$ \\
    Pre-Train on 2020 & $1.38596\times 10^{-3}$  & $1.46258\times 10^{-3}$ \\   
    Pre-Train on 2019 + 2020 & \textbf{1.38497} $\times \mathbf{10^{-3}}$  & \textbf{1.46254} $\times \mathbf{10^{-3}}$ \\ 
    \bottomrule
  \end{tabular}
  \caption*{\scriptsize Models pre-trained on Barcelona and Bangkok. Fine-Tuned on Moscow 2019, validated on Moscow 2020.}
  \label{table:pre-train}
\end{table}
\vspace{-4mm}
\subsubsection{Single Encoder vs Double Encoder}
Pre-training solves the issue that random initialisation of weights can hinder learning \cite{Erhan2010} and as such improves performance. However, for the task at hand any model developed must also be able to learn features specific to both 2019 and 2020, their similarities, differences and the ability to generalise between different cities. Despite the added performance of pre-training followed by fine tuning the act of continuing to train on top of these weights could lead to useful features diminishing or lost in favour of city specific features. It is due to this that a second encoder was added, $E_\phi$. The proposed model containing $E_\theta$, $D_\theta$ and now $E_\phi$ would still be pre-trained as before but prior to the fine-tuning step the weights in $E_\phi$ would be frozen. This allows $E_\theta$, $D_\theta$ and the skip-connections to learn city specific features on top of their pre-trained weights while $E_\phi$ learns a more generalised set of features that are city agnostic. This in essence allows $E_\theta$ and $E_\phi$ to act as two weak learners but when combined within the model architecture together, boosts performance.
\vspace{-4mm}
\begin{table}[H]
  \caption{Single vs Double Encoder}
  \centering
  \begin{tabular}{lll}
    \toprule
    \cmidrule(r){1-2}
    Type       & Training Loss &     Validation Loss \\ 
    \midrule
    Single Encoder & 3.01360 $\times 10^{-4}$  & 4.24992 $\times 10^{-4}$ \\ 
    Double Encoder & \textbf{3.00830} $\times \mathbf{10^{-4}}$  & \textbf{4.24517}  $\times \mathbf{10^{-4}}$ \\ \bottomrule
  \end{tabular}
  \caption*{\scriptsize Model pre-trained on Antwerp and Moscow. Fine-Tuned on Barcelona 2019, validated on Barcelona 2020.}
  \label{table:encoder}
\end{table}
\subsubsection{Spatial Dropout and 3D Convolution}
Model performance is further improved by adding Spatial Dropout  \cite{Tompson2015} after re-shaping and before a 1x1 3D Convolution layer. The role of Dropout is to reduce over-fitting caused by over-training; this is achieved by preventing activations from becoming strongly correlated. In regular dropout this is implemented by zeroing (``dropping out") the activations for that neuron. The problem with such an approach, especially in convolution environments is that adjacent pixels are likely to be highly correlated and zeroing activations may not help with reducing the dependency as much as originally intended hence it may be better to drop entire feature maps (Spatial Dropout). The 1x1 3D Convolution layer controls the number of feature maps and hence the number the of predicted frames, whether it be 12 for 5 minute granularity for the entire hour or just the 6 frames of interest.
\vspace{-4.5mm}
\begin{table}[H]
  \caption{Spatial Dropout + 3D Convolution}
  \label{sample-table}
  \centering
  \begin{tabular}{llll}
    \toprule
    & \multicolumn{2}{c}{Training Loss} \\
    \cmidrule(r){2-3}
    Type     & Pre-Train & Fine-Tune & Validation Loss \\
    \midrule
    Without Dropout & \textbf{4.50498} $\times \mathbf{10^{-4}}$   & 1.38406 $ \times 10^{-3}$   & 1.54008 $\times 10^{-3}$  \\
    With Dropout & 4.50755 $\times 10^{-4}$  & \textbf{1.37947} $\times \mathbf{10^{-3}}$  & \textbf{1.53572} $\times \mathbf{10^{-3}}$ \\
    \bottomrule
  \end{tabular}
  \caption*{\scriptsize Models pre-trained on Barcelona and Bangkok. Fine-Tuned on Moscow 2019, validated on Moscow 2020.}
  \label{table:spatial_dropout}
\end{table}
\vspace{-7mm}
\subsection{Training Process and Making Predictions}
The model described was trained using 4 Tesla V100s GPUs with a batch size of 4, learning rate of 1.5$\times10^{-3}$ with the LAMB optimiser \cite{You2019} (see Table \ref{table:lr} for learning optimiser comparisons). For the Core Challenge the model was pre-trained on all 4 additional cities using data from 2019 and 2020 for 15 epochs. Upon completion of the the pre-training $E_\phi$ was frozen and 4 copies of the model were made. These 4 pre-trained models were then each fine-tuned on a target city for 5 epochs resulting in a model for each target city of the core challenge.

\begin{figure}[H]
  \centering
  \begin{subfigure}[b]{0.3\textwidth}
    \includegraphics[width=\linewidth]{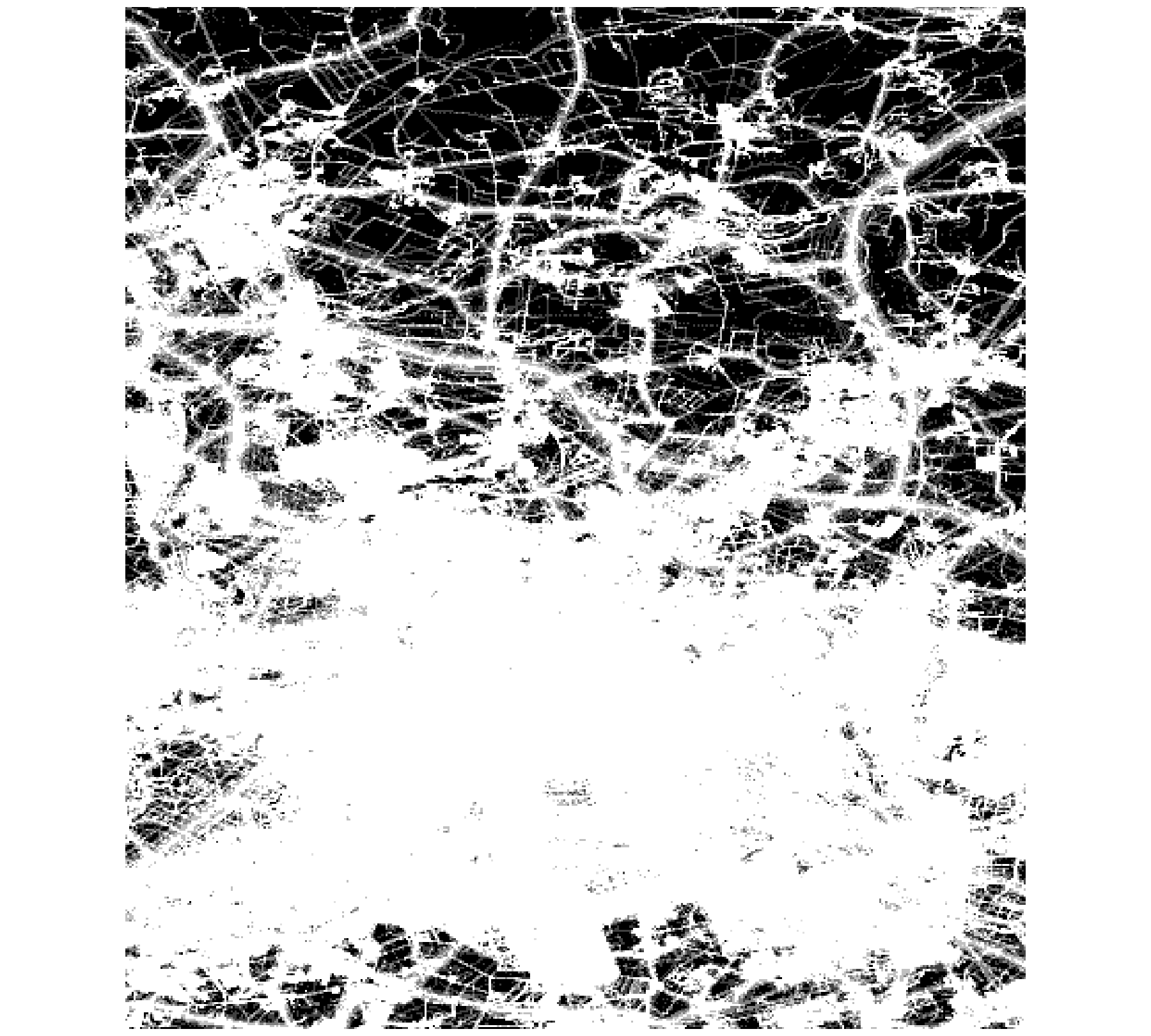}
    \caption{Mask from training data}
  \end{subfigure}
  \begin{subfigure}[b]{0.3\textwidth}
    \includegraphics[width=\linewidth]{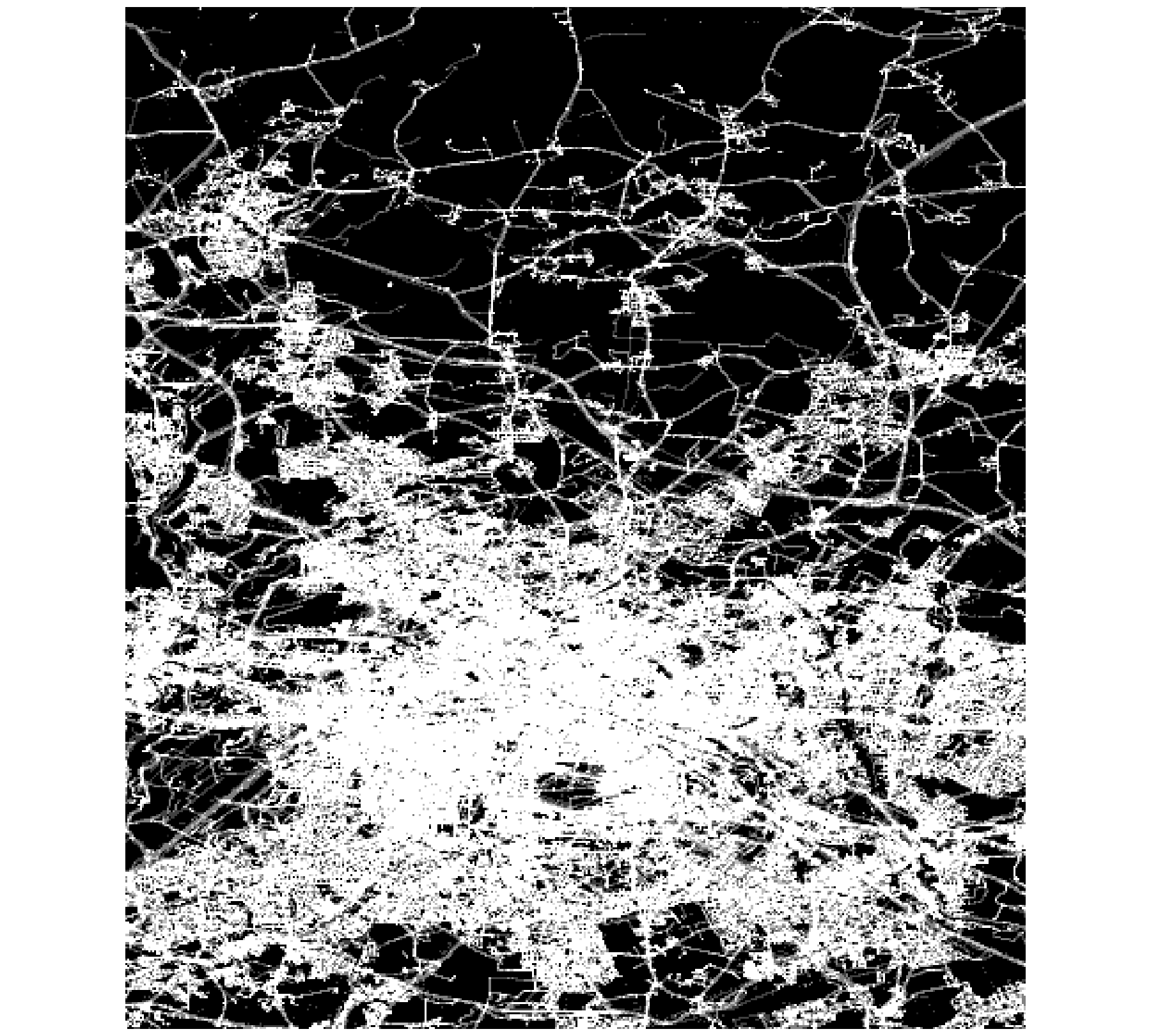}
    \caption{Mask from test data}
  \end{subfigure}
  \begin{subfigure}[b]{0.3\textwidth}
    \includegraphics[width=\textwidth]{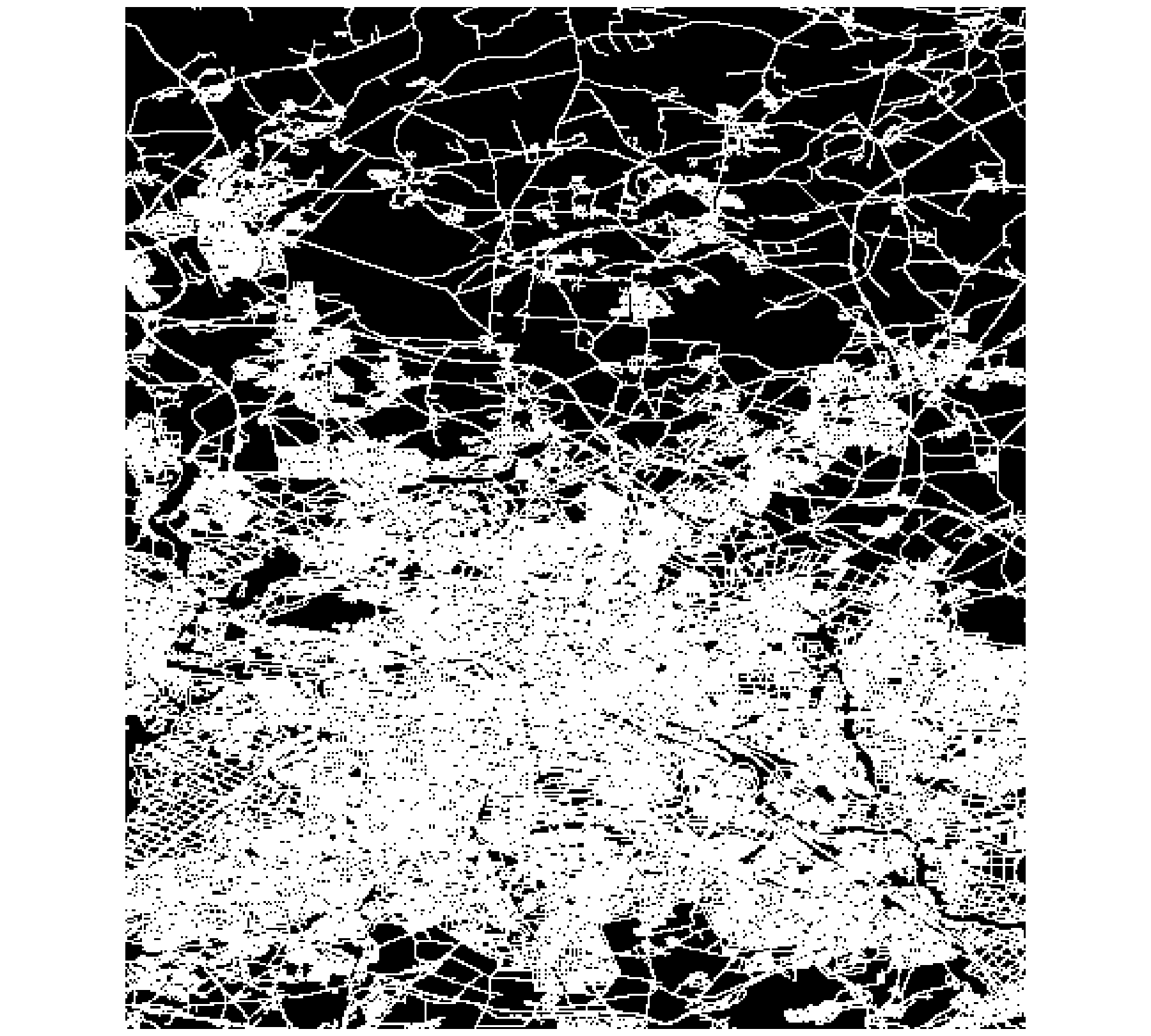}
    \caption{Mask from high res map}
  \end{subfigure}
  \caption{Berlin binary masks}
  \label{fig:masks}
\end{figure}

For the Extended Challenge the same training approach was adopted.
The mean of the 4 models' output was taken as the final predictions of the extended challenge cities. Interestingly it was observed that a smaller model achieved better results in the extended challenge than using the 4 models from the Core Challenge. This smaller model contained fewer filters and did not utilise Spatial Dropout or a 3D Convolution.
\vspace{-4.5mm}
\begin{table}[H]
  \caption{Learning Optimiser Selection}
  \centering
  \begin{tabular}{lll}
    \toprule
    \cmidrule(r){1-2}
    Type       & Training Loss &     Validation Loss \\
    \midrule
    SGD & $ 3.67464 \times 10^{-3}$  & $4.24061 \times 10^{-3}$ \\ 
    Adam & $ 1.38516 \times 10^{-3}$  &  $ 1.54384 \times 10^{-3}$ \\ 
    AdamW & $ 1.40625 \times 10^{-3}$  & $ 1.55830 \times 10^{-3}$ \\ 
    LAMB & $ \textbf{1.38406} \times \mathbf{10^{-3}}$  & \textbf{1.54008} $\times \mathbf{10^{-3}}$ \\ 
    \bottomrule
  \end{tabular}
  \caption*{\scriptsize Models pre-trained on Barcelona and Bangkok. Fine-Tuned on Moscow 2019, validated on Moscow 2020.}
  \label{table:lr}
\end{table}
\vspace{-6.5mm}
When making predictions, masking was also adopted. Three approaches to generating a mask were explored: using training data, using testing data and using the high resolution map - see Figure \ref{fig:masks}. Masks generated using the testing data helped obtain the best predictions.
\vspace{-1.5mm}


\section{Discussion}
The model and approach described throughout this paper has led to a competitive competition score in both the core and extended challenge, achieving 50.6890 and 60.8776 respectively. 

One of the main highlights that makes this model stand out advantageous over potential competitors is the number of trainable parameters. The 4 models used for the core challenge had roughly 460k parameters per model and the 4 models used for the extended challenge had roughly 120k parameters per model. By comparison the 1st and 2nd place finishers of Traffic4cast 2020 \cite{Choi2020, Wu2020} utilised 10s of Convolution Blocks each containing multiple convolution layers resulting in millions of parameters whereas the approach described in this paper uses only 9 Convolutional LSTM layers, 1 3D Convolution layer and 2 Transposed Convolution layers resulting in significantly fewer parameters.

Given this smaller model size and competitive performance should real-world model deployment be considered, where real-time response is more important, it would be much more advantageous than a larger model that only gives small improvements. With the nature of mobility and the continuously changing environment, models may need to be re-trained at regular intervals (a period of months) and compromise between training time, inference time and model performance will need to be made. A lightweight model with faster training and inference times that achieves nearly as good of a performance as a larger heavy-weight model with significantly longer training and inference times will always be more beneficial.

\section{Conclusion and Future Work}
This paper has presented a Dual Encoding U-Net for the purpose of Traffic Forecasting presented as a frame prediction task created by IARAI's Traffic4cast 2021. The model and approach described achieves good performance in predicting 2020 traffic behaviour of certain cities when only 2019 training data is available as well as making predictions on unknown cities where no training data is available - obtaining competitive competition MSE scores of 50.6890 and 60.8776 in the Core and Extended challenges respectively.

A notable contribution from this work includes the introduction of two forms of skip-connections that are more appropriate when using Convolutional LSTM layers rather than maintaining the same approach used between standard convolution layers. In addition to this, the model architecture and training methodology proposed contains significantly fewer parameters and utilises shorter training times than many Deep Learning competitors in traffic forecasting. Thus making it an extremely desireable approach when looking at potential practical applications with real world deployment. 

An interesting area for future research could be a variational encoder married with the current model. Such an approach was attempted where $E_\phi$ was variational producing a mean and variance to describe a distribution, however due to consistent collapse in Kullback-Leibler loss and difficulty training it was sidelined. With more time and experimentation it is felt that such an approach could be beneficial; a similar approach has proved successful in weather frame prediction \cite{Qi} and thus could also prove effective for traffic frame prediction. 




\section*{About Us}
Alchera Data Technologies Ltd has developed a cloud-based AI software, Alpha, powering enterprise-grade intelligent mobility and intelligent infrastructure applications.

It provides operators and commercial users of road infrastructure with absolute, real-time data on vehicle and pedestrian movements around cities and major infrastructure at extreme scale.

Alpha operates with greater coverage, lower cost and with greater reliability than any existing solutions, by fusing together existing sensor networks (e.g., CCTV, IoT, mobile, connected car, etc.) to build an enriched comprehensive single data feed and delivering first-of-a-kind City Scale, Real-Time Digital Twins.

\clearpage
\bibliographystyle{IEEEtran}
\bibliography{main.bib}

\end{document}